\documentclass[letterpaper, 10 pt, conference]{ieeeconf}  

\IEEEoverridecommandlockouts                              

\usepackage{graphics} 
\usepackage{epsfig} 
\usepackage{mathptmx} 
\usepackage{times} 
\usepackage{amsmath} 
\usepackage{amssymb}  
\usepackage[T1]{fontenc}
\usepackage{todonotes}
\usepackage{color,soul}
\usepackage{hyperref}
\usepackage{cleveref}

\newcommand\copyrighttext{%
  \footnotesize \textcopyright 2024 IEEE. Personal use of this material is permitted.
  Permission from IEEE must be obtained for all other uses, in any current or future
  media, including reprinting/republishing this material for advertising or promotional
  purposes, creating new collective works, for resale or redistribution to servers or
  lists, or reuse of any copyrighted component of this work in other works. Presented at ICRA 2024 in Yokohama, Japan.
  DOI: \href{https://doi.org/10.1109/ICRA57147.2024.10610662}{10.1109/ICRA57147.2024.10610662}}
\newcommand\copyrightnotice{%
\begin{tikzpicture}[remember picture,overlay]
\node[anchor=south,yshift=10pt] at (current page.south) {\fbox{\parbox{\dimexpr\textwidth-\fboxsep-\fboxrule\relax}{\copyrighttext}}};
\end{tikzpicture}%
}

\title{\LARGE \bf
A novel metric for detecting quadrotor loss-of-control* 
}

\author{Jasper J. van Beers$^{1}$, Prashant Solanki$^{2}$, and Coen C. de Visser$^{3}$
\thanks{*This work is supported through The Dutch Research Council (NWO) VIDI grant awarded to dr. Coen de Visser}
\thanks{$^{1}$Jasper van Beers is with the Faculty of Aerospace Engineering, Delft University of Technology, 2629 HS Delft, The Netherlands
        {\tt\small j.j.vanbeers@tudelft.nl}}%
\thanks{$^{1}$Prashant Solanki is with the Faculty of Aerospace Engineering, Delft University of Technology, 2629 HS Delft, The Netherlands
        {\tt\small p.solanki@tudelft.nl}}%
\thanks{$^{3}$dr. Coen de Visser is with the Faculty of Aerospace Engineering, Delft University of Technology, 2629 HS Delft, The Netherlands
        {\tt\small c.c.devisser@tudelft.nl}}%
}

\begin{document}

\maketitle
\copyrightnotice
\thispagestyle{empty}
\pagestyle{empty}

\begin{abstract}

Unmanned aerial vehicles (UAVs) are becoming an integral part of both industry and society. In particular, the quadrotor is now invaluable across a plethora of fields and recent developments, such as the inclusion of aerial manipulators, only extends their versatility. As UAVs become more widespread, preventing loss-of-control (LOC) is an ever growing concern. Unfortunately, LOC is not clearly defined for quadrotors, or indeed, many other autonomous systems. Moreover, any existing definitions are often incomplete and restrictive. A novel metric, based on actuator capabilities, is introduced to detect LOC in quadrotors. The potential of this metric for LOC detection is demonstrated through both simulated and real quadrotor flight data. It is able to detect LOC induced by actuator faults without explicit knowledge of the occurrence and nature of the failure. The proposed metric is also sensitive enough to detect LOC in more nuanced cases, where the quadrotor remains undamaged but nevertheless losses control through an aggressive yawing manoeuvre. As the metric depends only on system and actuator models, it is sufficiently general to be applied to other systems.

\end{abstract}

\section{INTRODUCTION}

\subsection{Motivation}
As the world becomes more automated, and various robots become more integrated with daily life, the ability of these systems to recognize and prevent accidents is essential to their widespread and accepted use. In fact, Carlson et al. \cite{RobotFailure2004} found that the main cause of failure in many robotic applications is the control system (henceforth referred to as loss-of-control, LOC), followed by mechanical failures (e.g. actuators). Moreover, studies which find that the increase in automation and application of robotics often lead to more accidents, such as for aircraft autopilots \cite{OLIVER2019772} and manufacturing robots \cite{YANG2022105623}, only underscore this need for better safety. 

In recent years, unmanned aerial vehicles (UAVs) have enjoyed a surge of popularity and use in the field of robotics and beyond. UAVs have become indispensable in a plethora of applications, such as: agriculture \cite{app12031047}, (potentially hazardous) maintenance \& inspection \cite{drones6060137}, medical operations \cite{nisingizwe2022effect}, and videography. Recent literature on UAVs extends their versatility by equipping them with manipulators \cite{Hamaza2018_AerialManipulator}. These manipulators may facilitate repairs of infrastructure, such as off-shore wind turbines \cite{drones6060137}, or improve operational efficiency by enabling perching \cite{zheng2023metamorphic}.

Given this explosion of research, and the perpetual concern of safety, it comes as no surprise that an extensive body of research focuses on improving UAV safety (e.g. \cite{Sun2019_MonteCarlo,sun2021autonomous,Schieni_QuadrotorFEP,RNN_LOC_Altena}). In particular, many of the advancements are centered around the under actuated - and thus vulnerable - quadrotor, which is perhaps the most popular variant of the UAV. However, many of these works (e.g. \cite{sun2021autonomous,WANG2020105745,NeuroAdaptiveFTC_Song2019,Zogopoulos2021_FTC_FixedWingUAV_FE_awareness,MPC_FTC_Nan2022}) are concerned with fault-tolerant control (FTC), and thus do not directly address cases where the undamaged quadrotor experiences loss-of-control (LOC). Furthermore, LOC itself is not clearly defined for the quadrotor; the few LOC definitions that exist are based on simple attitude constraints and are thus too restrictive \cite{Schieni_QuadrotorFEP,RNN_LOC_Altena}. Therefore, to encourage the advancement of LOC-aware quadrotors (and by extension other autonomous systems), there is a need to properly detect and label moments of LOC.



\subsection{Related Work}


With regards to FTC, many derive unique controller architectures to accommodate hovering and high-speed flight of the quadrotor, despite single and double rotor loss by forfeiting yaw control \cite{sun2021autonomous,Sun_ControlDoubleFailure,sun2018high}. Other FTC schemes include those based on sliding mode control \cite{WANG2020105745,FTC_SlidingMode_Sharifi2010}, machine learning \cite{NeuroAdaptiveFTC_Song2019}, and model predictive control \cite{MPC_FTC_Nan2022}. 


In aircraft literature, LOC is well-studied and is often defined as an excursion from the safe flight envelope (SFE) bounds \cite{international2015loss,Barlow2011_EstimatingLOC_DataDriven,Wilborn2004_LOCDefinitionAircraft,Rafi2021_Realtime_LOC_mitigation}. Due to the curse of dimensionality, obtaining estimates of the SFE is difficult and computationally intractable through standard Hamilton-Jacobi reachability analyses \cite{Zhidong2022_FEP_Reachability}. Therefore, Sun et al. \cite{Sun2019_MonteCarlo} employ Monte Carlo simulations to approximate a 6-state SFE of a quadrotor through optimal (bang-bang) control. However, this approach does not accommodate the use of other controllers. Instead, other efficient SFE estimation techniques, such as \cite{Zhidong2022_FEP_Reachability}, may be used. However, any estimated SFEs emerge from a given (nominal) model of the system and thus do not directly account for changing dynamics and/or faults. Deriving a family of, already computationally intensive, SFEs can be impractical.



Others instead define quadrotor LOC based on attitude constraints. For example, Schieni et al. \cite{Schieni_QuadrotorFEP} hedge the allowable attitude of the quadrotor to limit its aggressiveness, thereby avoiding dangerous manoeuvres. Altena et al. \cite{RNN_LOC_Altena} train a recurrent neural network to predict quadrotor LOC using flight data of an undamaged quadrotor where LOC is defined as the moment when the roll or pitch angle exceed 90 deg. This is tailored to the LOC scenario of the data set. These definitions are prohibitive and scenario specific.

\subsection{Contributions}

The main contribution of this paper is a novel indicator, termed the \textit{feasibly controllable metric} (FCM), that defines LOC for quadrotors. By applying the proposed LOC detector to both simulated and real flight data, we demonstrate that the FCM is not only sensitive to LOC detection under actuator faults, but is also capable of detecting LOC when no faults are present (i.e. in the undamaged case). 

Moreover, since the FCM derives from system and actuator models, it is sufficiently general to be compatible with other robotic or autonomous platforms where such state measurements and suitable actuator models are available.  

\subsection{Organization}
The remainder of this paper is organized as follows: \cref{sec:quadrotor_background} briefly describes the quadrotor equations of motion and aerodynamic models used in this study. The FCM metric itself is derived in \cref{sec:method} and applied to simulation and flight data in \cref{sec:results}.

\section{QUADROTOR MODEL}\label{sec:quadrotor_background}
The equations of motion for the quadrotor, expressed in the body reference frame, may be described by
\begin{equation}\label{eq:QuadSimple_Force_B}
    m \left ( \dot{\mathbf{V}}_{B} + \boldsymbol{\Omega}_{B} \times \mathbf{V}_{B} \right ) = R_{BE}m\mathbf{g} + \mathbf{F}_{B}
\end{equation}

\begin{equation}\label{eq:QuadSimple_Moment_B}
    \mathbf{I_{v}}\dot{\boldsymbol{\Omega}}_{B} + \boldsymbol{\Omega}_{B} \times \mathbf{I_{v}}\boldsymbol{\Omega}_{B} = \mathbf{M}_{B}
\end{equation}
where, the mass and moment of inertia of the quadrotor are denoted by $m$ and $I_{v}$ respectively. The gravity vector, defined in the inertial (North-East-Down, NED) reference frame is given by $\mathbf{g}$ and the rotation matrix transforming a vector from the inertial frame to the body frame is represented by $R_{EB}$. The body translational and rotational velocities are described by $\mathbf{V}_{B} = [u, v, w]^{T}$ and $\boldsymbol{\Omega}_{B} = [p, q, r]^{T}$ respectively. $\mathbf{F}_{B} = [F_{x}, F_{y}, F_{z}]^{T}$ and $\mathbf{M}_{B} = [M_{x}, M_{y}, M_{z}]^{T}$ represent the (aerodynamic) forces and moments acting on the quadrotor body. 

The non-linear $\mathbf{F}_{B}$ and $\mathbf{M}_{B}$ can be identified from flight data through step-wise regression \cite{Klein2006_AircraftSysID}. Indeed, step-wise regression has previously, and successfully, been applied and numerically validated for quadrotors (interested readers are directed to \cite{Sun2019_unDamagedIDQuadHS,Sun2018_GrayBox,PI_Validation_vanBeers} for more details on the identification procedure and model structure selection). We here employ the models of \cite{PI_Validation_vanBeers}, for which the underlying high-speed flight data is collected (at 500Hz) via free flights in a wind tunnel with wind speeds ranging from 0-14 m/s.


While the LOC detection method proposed in \cref{sec:method} should be generalizable to other controlled systems where state and actuator models/measurements are available, we here apply the metric to the (inner-loop) rate control of a quadrotor. Rearranging \cref{eq:QuadSimple_Moment_B} and replacing $\mathbf{M_{B}}$ with the appropriate identified polynomial aerodynamic models of \cite{PI_Validation_vanBeers} yields \cref{eq:momentRearranged}.


\begin{equation}\label{eq:momentRearranged}
    \dot{\boldsymbol{\Omega}}_{B} = \mathbf{I_{v}}^{-1}\left [ \begin{array}{c}
         M_{x}(p, u_{p}) \\ 
         M_{y}(q, u_{q}) \\
         M_{z}(r, u_{r}) \\
    \end{array} \right ]
    - \mathbf{I_{v}}^{-1}\boldsymbol{\Omega}_{B} \times \mathbf{I_{v}}\boldsymbol{\Omega}_{B}
\end{equation}
In \cref{eq:momentRearranged}, $u_{p}$, $u_{q}$, and $u_{r}$ represent the difference in rotor speeds that enact rolling, pitching, and yawing moments respectively. This is used as the input vector in this paper since it results in a diagonal control effectiveness matrix, $B$, in \cref{sec:method}. The control moments depend on the quadrotor configuration and, for the quadrotors used in this paper\footnote{$\omega_{1}$ is located front left, $\omega_{2}$ front right, $\omega_{3}$ aft right and $\omega_{4}$ aft left.}, may be constructed from the rotor speeds via
\begin{equation}
    \begin{array}{cl}
        u_{p} = & \left ( {\omega_{1}} + {\omega_{4}} \right ) - \left ( {\omega_{2}} + {\omega_{3}} \right ) \\
        u_{q} = & \left ( {\omega_{1}} + {\omega_{2}} \right ) - \left ( {\omega_{3}} + {\omega_{4}} \right ) \\
        u_{r} = & S_{r} \left [ \left ( {\omega_{1}} + {\omega_{3}} \right ) - \left ( {\omega_{2}} + {\omega_{4}} \right ) \right ]
    \end{array}
\end{equation}
 $S_{r} = 1$ for clockwise rotation of $\omega_{1}$ and $S_{r} = -1$ if counter-clockwise. The aerodynamic moment models themselves are (up to) forth order polynomial functions of the associated body rotational rates $\boldsymbol{\Omega_{B}}$ and control moments \cite{Sun2018_GrayBox,PI_Validation_vanBeers}. 
 

\section{DERIVATION OF FEASIBLY CONTROLLABLE METRIC}\label{sec:method}
Consider a general non-linear system described by 
\begin{equation}\label{eq:general_nonlinear}
    \mathbf{\dot{x}} = f(\mathbf{x},\mathbf{u})
\end{equation}
where $\mathbf{x} \in \mathbb{X} \subset \mathbb{R}^{n}$ denotes the system states and $\mathbf{u} \in \mathbb{U} \subset \mathbb{R}^{m}$ describes the inputs of the system.

The difference of \cref{eq:general_nonlinear} between two consecutive samples, $k$ and $k+1$, may be approximated (assuming, $f(\mathbf{x},\mathbf{u})$ is smooth and differentiable) through
\begin{equation}\label{eq:difference}
\begin{array}{rll}
    \mathbf{\dot{x}}_{k+1} - \mathbf{\dot{x}}_{k} & = &  \frac{\delta f}{\delta \mathbf{x}} \cdot (\mathbf{x}_{k + 1} - \mathbf{x}_{k}) + \frac{\delta f}{\delta \mathbf{u}} \cdot (\mathbf{u}_{k + 1} - \mathbf{u}_{k}) \\
    \to d\mathbf{\dot{x}} & = & \frac{\delta f}{\delta \mathbf{x}} d\mathbf{x} + \frac{\delta f}{\delta \mathbf{u}} d\mathbf{u}
\end{array}
\end{equation}

Taking the time derivative of this difference yields
\begin{equation}
\begin{array}{rll}
    \frac{1}{dt} \cdot \left ( d\mathbf{\dot{x}} \right )& = & \frac{1}{dt} \cdot \left (\frac{\delta f}{\delta \mathbf{x}} d\mathbf{x} + \frac{\delta f}{\delta \mathbf{u}} d\mathbf{u} \right ) \\
    \to \mathbf{\ddot{x}} & = & \frac{\delta f}{\delta \mathbf{x}} \mathbf{\dot{x}} + \frac{\delta f}{\delta \mathbf{u}} \mathbf{\dot{u}}
\end{array}
\end{equation}
\Cref{eq:difference} is only truly valid if $d\mathbf{x}$ and $d\mathbf{u}$ are small. For dynamic systems such as the quadrotor, this holds if the time increment, $dt$, is sufficiently small\footnote{Many open-source flight controllers, such as BetaFlight, support sampling rates of up to 8kHz.}. Hence, taking the time derivative is an elegant way to linearize the system - with respect to its derivative - to arrive at the so-called differential form (or, alternatively, the velocity form \cite{VelocityForm_Leith1999}):
\begin{equation}\label{eq:differential_form}
    \mathbf{\ddot{x}} = A \mathbf{\dot{x}} + B \mathbf{\dot{u}}
\end{equation}

\Cref{eq:differential_form} gives an exact representation of the non-linear system at the point $(\mathbf{x}, \mathbf{u})$. In other words, it may be considered a linear parameter varying (LPV) model, scheduled by the states and inputs. For the fortunate cases where $f(\mathbf{x},\mathbf{u})$ is known (and differentiable), relations for $A$ and $B$ may be exactly derived. Even if $f(\mathbf{x},\mathbf{u})$ is known, linearization is nonetheless useful as it facilitates the application of powerful and well-studied linear analysis tools. While not strictly necessary for the method described here, the differential form offers a simple way to derive a valid linear formulation, in particular for the control effectiveness $B$, of the system.


Often, the standard definition for controllability (i.e. \cref{eq:controllability}) is too lenient as it does not consider the limits of the actuator. Indeed, for a system to be \textit{feasibly controllable}, the actuators should be capable of enacting the change necessary to maintain control. This may not always be the case, especially in the presence of partial actuator faults or saturation. However, purely looking at actuator saturation for LOC detection is too restrictive as it may be necessary to (at least temporarily) saturate the actuators to satisfy reference tracking. Consider, for example, a "punch out" manoeuvre for the quadrotor or other such aggressive maneuvering which often saturates the rotors, but are often not considered LOC events\footnote{Occasionally, however, these aggressive manoeuvres can lead to LOC, suggesting that saturation may play a crucial role in LOC detection.}. The well-known controllability matrix is: 

\begin{equation}\label{eq:controllability}
    \begin{array}{ccccccc}
        C & = & [B & AB & A^{2} B & \cdots & A^{n-1} B]
    \end{array}
\end{equation}
In general, $f(\mathbf{x},\mathbf{u})$ is mostly unknown and thus $A$ and $B$ need to be estimated. These may be estimated directly through \cref{eq:differential_form}, or derived using the velocity form applied to a full non-linear model that approximates $f(\mathbf{x}, \mathbf{u})$ identified through other techniques, such as neural networks or step-wise regression \cite{Klein2006_AircraftSysID,Sun2018_GrayBox}.

Note that, for fast systems such as the quadrotor, the time increment between samples is often small and thus $d\mathbf{x} \approx 0$. While not strictly necessary for the metric, we will make the approximation that $d\mathbf{\dot{x}} \approx B \cdot d\mathbf{u}$ when dealing with real flight data in \cref{sec:results} to address cases where reliable estimates of $A$ may be unavailable. The $d\mathbf{\dot{x}} \approx B \cdot d\mathbf{u}$ approximation can be made for systems with similar (actuator) characteristics, but should not be made when $A\cdot d\mathbf{x}$ is not negligible across consecuetive time steps. 


However, unless $B$ (and equivalently, $A$ if available) is estimated online, the identified models typically refer to the nominal system and thus may not account for time-varying dynamics or actuator faults. Indeed, the ability of a system to maintain control depends on its current dynamics and actuator capabilities. We therefore suggest that $A$ and $B$ be locally updated in some capacity. The cross-correlation\footnote{For the quadrotor, $M < 0$ should not occur as the control effectiveness associated with the control moments cannot switch. Moreover, $M=0$ is avoided as this leads to a singular $B$.}, $M \in (0, 1]$, between $d\mathbf{\dot{x}}$ and $A\cdot d\mathbf{x} + B\cdot d\mathbf{u}$ may be used as an indicator of the discrepancy between the nominal and current systems: any changes in the state derivative should be explainable by a corresponding change in the (one-step behind) inputs. The nominal control effectiveness may then be modified through a product with $M$ (i.e. $B_{k+1} = B_{k}M$) where $M$ is estimated over a moving window. Other methods for obtaining $M$, such as Recursive Least Squares, may also be used for better estimates. 

The core principle behind the proposed metric is that, under perfect conditions, changes in the state derivative are fully explainable by the model. Thus, 
\begin{equation}\label{eq:nominal_difference}
    d\mathbf{\dot{x}} - A\cdot d\mathbf{x} - B \cdot (\mathbf{u_{k+1}} - \mathbf{u_{k}}) = 0
\end{equation}

However, errors - be these from model misspecification, a change in dynamics, or faults - will likely be observed. These should be accommodated by an additional control effort as shown in \cref{eq:FCMetric_desired_corrective_control}. Note that $\mathbf{e}$ here refers to the error between measurements of the state derivative and predictions from the (difference) model (\cref{eq:difference}), and not the tracking error. 
\begin{equation}\label{eq:FCMetric_desired_corrective_control}
    d\mathbf{\dot{x}} - A\cdot d\mathbf{x} - B \cdot (\mathbf{u_{k+1}} - \mathbf{u_{k}} + \Delta\mathbf{u_{c}}) = \mathbf{e}
\end{equation}

Substituting \cref{eq:nominal_difference} into \cref{eq:FCMetric_desired_corrective_control} and solving for the necessary corrective control action yields
\begin{equation}
    \Delta\mathbf{u_{c}} = B^{-1}\mathbf{e}
\end{equation}

The key, then, is to determine if such a corrective command is possible, and if not, we consider the system to be in a state of instantaneous LOC. Indeed, the allowable periods of LOC may differ between systems and operating conditions. Therefore, a majority voting window - tied to a specified tolerable duration of sustained LOC - may be employed to identify periods of LOC. These may be related to the time-to-recovery, or backward reachable set, of the system.

An actuator model may be used to probe the feasibility of a particular command. We here rely on a first order actuator model of the quadrotor, however, other (potentially non-linear) actuator models are compatible with the proposed approach so long as the maximum actuator effort can be found at each time step. The maximum possible change in the control moments prescribed by the first order motor model is given by 
\begin{equation}\label{eq:quad_act_first_order}
    \Delta\mathbf{u_{MAX}} = dt \cdot \sum_{i=1}^{4}  \frac{1}{\tau}\left ( \omega_{MAX} - \omega_{i} \right ) 
\end{equation}
where $dt$ denotes the time step, $\tau$ the actuator time constant (which can be estimated from flight data), and $\omega_{MAX}$ is the lower/upper saturation limits of the rotor. Whether the lower or upper limit should be used depends on the direction of necessary change in rotor speed in order to correct for $\mathbf{e}$. 

The quadrotor is considered to be in a state of LOC if $\Delta\mathbf{u_{MAX}} < \Delta\mathbf{u_{c}}$, even when \cref{eq:controllability} maintains full row rank. We define the \textit{Feasibly Controllable Metric} (FCM), at an instant, $k$, to be
\begin{equation}\label{eq:FCM_k}
    FCM[k] = \left \{ 
    \begin{array}{cr}
        0 & \text{if } \Delta\mathbf{u_{MAX}}[k] < \Delta\mathbf{u_{c}}[k] \text{ or } \text{rank}(\mathbf{C}[k]) < n\\ 
        1 & \text{otherwise} 
    \end{array} \right .
\end{equation}
This may be extended to a majority voting window by taking the average of $FCM[k]$ across MVW samples to yield the windowed FCM:
\begin{equation}\label{eq:FCM_moving}
    FCMW = \left \{ 
    \begin{array}{cr}
        0 & \text{if } \frac{1}{MVW}\sum_{i=k-MVW}^{k} FCM[i] < 0.5
        \\
        1 & \text{otherwise} 
    \end{array} \right .
\end{equation}

Note that, for many systems such as the quadrotor, direct measurements of $\dot{x}$ (and $\ddot{x}$) are usually unavailable. These therefore need to be estimated (from the state), which often introduces a lag between knowledge of the current state and corresponding derivative. Thus, the state, its derivative, and input vectors should be properly synchronized when computing the FCM through \cref{eq:differential_form}, \cref{eq:FCMetric_desired_corrective_control} and \cref{eq:quad_act_first_order}. Subsequently, the FCM information is also delayed with respect to the current index by a few samples, which can easily be compensated for in (a-posteriori) analyses by shifting the time-series. In online applications, this delay in FCM delivery is negligible if LOC manifests on a relatively longer time-scale compared to the sampling rate (i.e. time-to-LOC $\gg$ time step). 

\section{RESULTS: SIMULATION AND FLIGHT DATA}\label{sec:results}


We first apply the $FCMW$ metric on a simulated quadrotor to demonstrate 
its use for cases with full model knowledge and actuator faults. Such faults are considered true LOC scenarios. As a more practical demonstration, the $FCMW$ is also applied to real quadrotor flight data where LOC is suspected, but not guaranteed, to occur. For this case, a full model of the quadrotor is assumed to be unavailable, and thus the $FCMW$ is derived only from estimates of $B$ made on the flight data. Note, however, that the implicit assumption made here is that the quadrotor remains (theoretically) controllable by $rank(\mathbf{C}) = n$. This assumption is reasonable since no (partial) actuator faults occur in the flight data.  

\subsection{Simulation Overview}

The identified high-speed aerodynamic models of \cite{PI_Validation_vanBeers} are deployed in a Python-based non-linear quadrotor simulation environment running at 250 Hz. Control of the quadrotor is facilitated through cascaded PID loops which map desired position and yaw commands to desired accelerations and rates. The inner-most rate and acceleration loop utilize the incremental non-linear dynamic inversion (INDI) controller developed by \cite{SMEUR201879_INDI}. Briefly, the goal of the controller is to align the thrust-vector of the quadrotor with the desired acceleration vector in the NED frame. The simulated quadrotor's properties are summarized in \cref{tab:HDBeetle_properities}. 

\begin{table}[!b]
\caption{Physical properties of the simulated quadrotor}
\label{tab:HDBeetle_properities}
\centering
\begin{tabular}{lr} \hline
\multicolumn{1}{l}{\textbf{Mass (incl. batteries)}, $g$}            & 433           \\
\multicolumn{1}{l}{\textbf{Diagonal hub-to-hub}}                    & 217           \\
\multicolumn{1}{l}{\textbf{  diameter}, $mm$}                         & \\
\multicolumn{1}{l}{\textbf{Propeller diameter}, $mm$}               & 76.5          \\
\multicolumn{1}{l}{\textbf{Moment of inertia}, $kg\cdot m^{2}$}     & $10^{-3} \cdot \left [ \begin{array}{ccc}
    0.865 & 0 & 0 \\
    0 & 1.07 & 0 \\
    0 & 0 & 1.71
\end{array} \right ]$ \\ 
\multicolumn{1}{l}{\textbf{Actuator time constant}, $Hz$} & 60   \\
\hline
\end{tabular}
\end{table}

Furthermore, zero-mean white noise is injected into the simulated measurements of the quadrotor's velocity, $\mathcal{N}(0, \sigma_{V} = 0.01 )$m/s, and rotational rates, $\mathcal{N}(0, \sigma_{\Omega} = 8.73\cdot 10^{-5})$rad/s. These values are derived from the inertial measurement unit (MPU6000) of the quadrotor used to generate the underlying aerodynamic models for the simulator. 

\subsection{Loss-of-control Detection in Simulation}\label{subsec:simRes}

\begin{figure}[!b]
    \centering
    \includegraphics[width = \columnwidth]{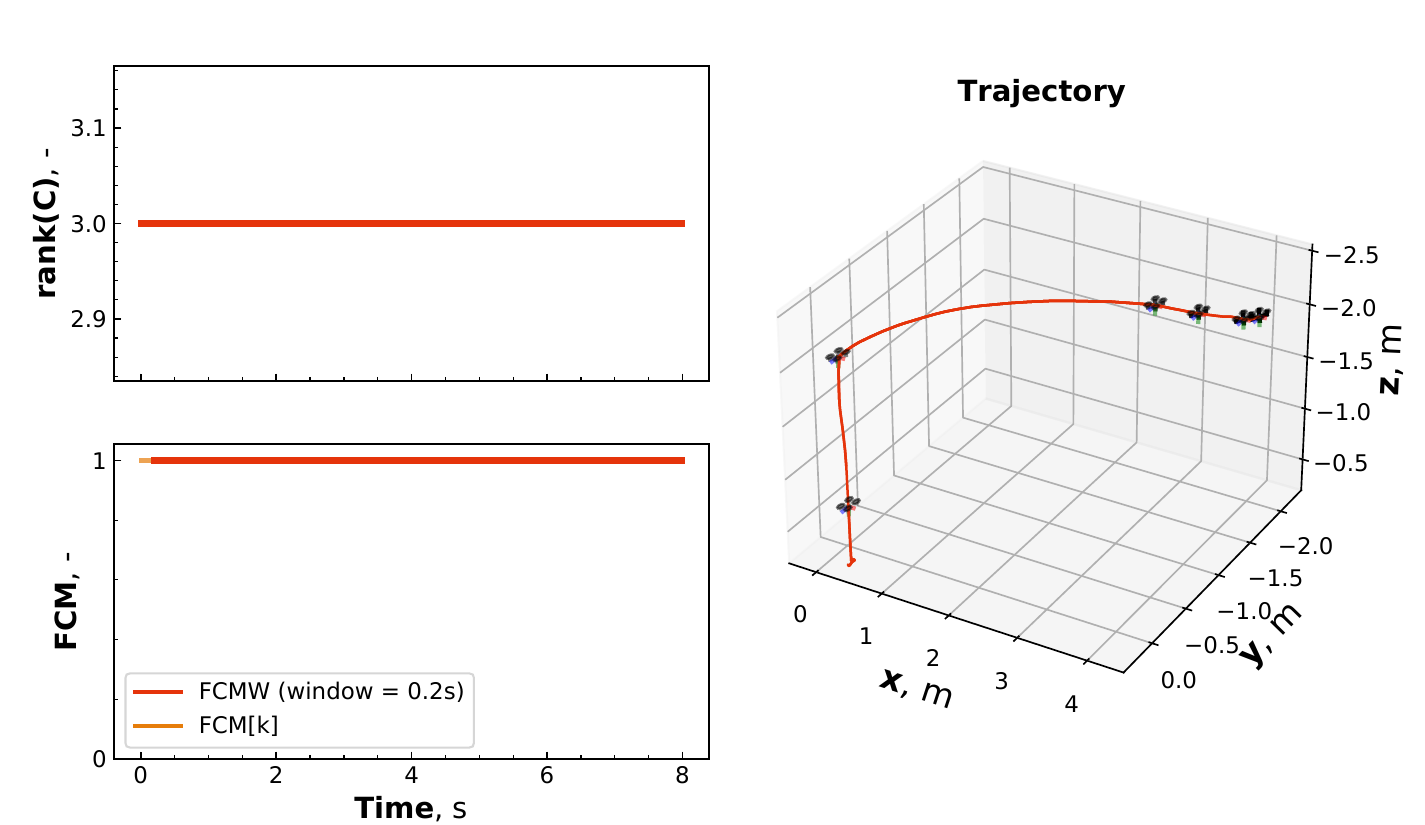}
    \caption{Position tracking trajectory of the nominal quadrotor. Also shown are the standard controllability metric, $rank(C) = n$, and those proposed in this paper, $FCM[k]$ and $FCMW$.}
    \label{fig:PosTrack_NOM}
\end{figure}

Using the simulator, two experiments are conducted wherein the quadrotor is assigned a simple position tracking task. In the first experiment, the quadrotor is in its nominal configuration and experiences no actuator faults. In this scenario, the proposed metrics are not expected to detect LOC. Conversely, in the second experiment, one of the rotors is capped at 0\% thrust at $t=5.2$s, representing the complete loss of a single rotor. Here, the proposed metrics are expected to be able to detect LOC. In this experiment, the metrics are not explicitly made aware of the fault (i.e. \cref{eq:quad_act_first_order} assumes that all rotors are still fully operational).

In both experiments, the current control effectiveness matrix is updated through $B_{k+1} = B_{k}M$ as described in \cref{sec:method}. Here, a moving window of $0.2$s is used to determine the values of $M$ where, nominally, $M=I$. Recall that $B$ (and thus $M$) are diagonal as the quadrotor control scheme employed here relies on the (orthogonal) control moments as the input vector. Furthermore, the majority voting window duration for the $FCMW$ is also set to $0.2$s to mirror $M$.

\Cref{fig:PosTrack_NOM} depicts the position tracking trajectory of the quadrotor in its nominal configuration (i.e. no actuator faults). As expected, $rank(C) = 3$ and both the $FCM[k] = 1$ and $FCMW = 1$ suggesting that the quadrotor remains in control for the duration of the flight.

\begin{figure}[!b]
    \centering
    \includegraphics[width = \columnwidth]{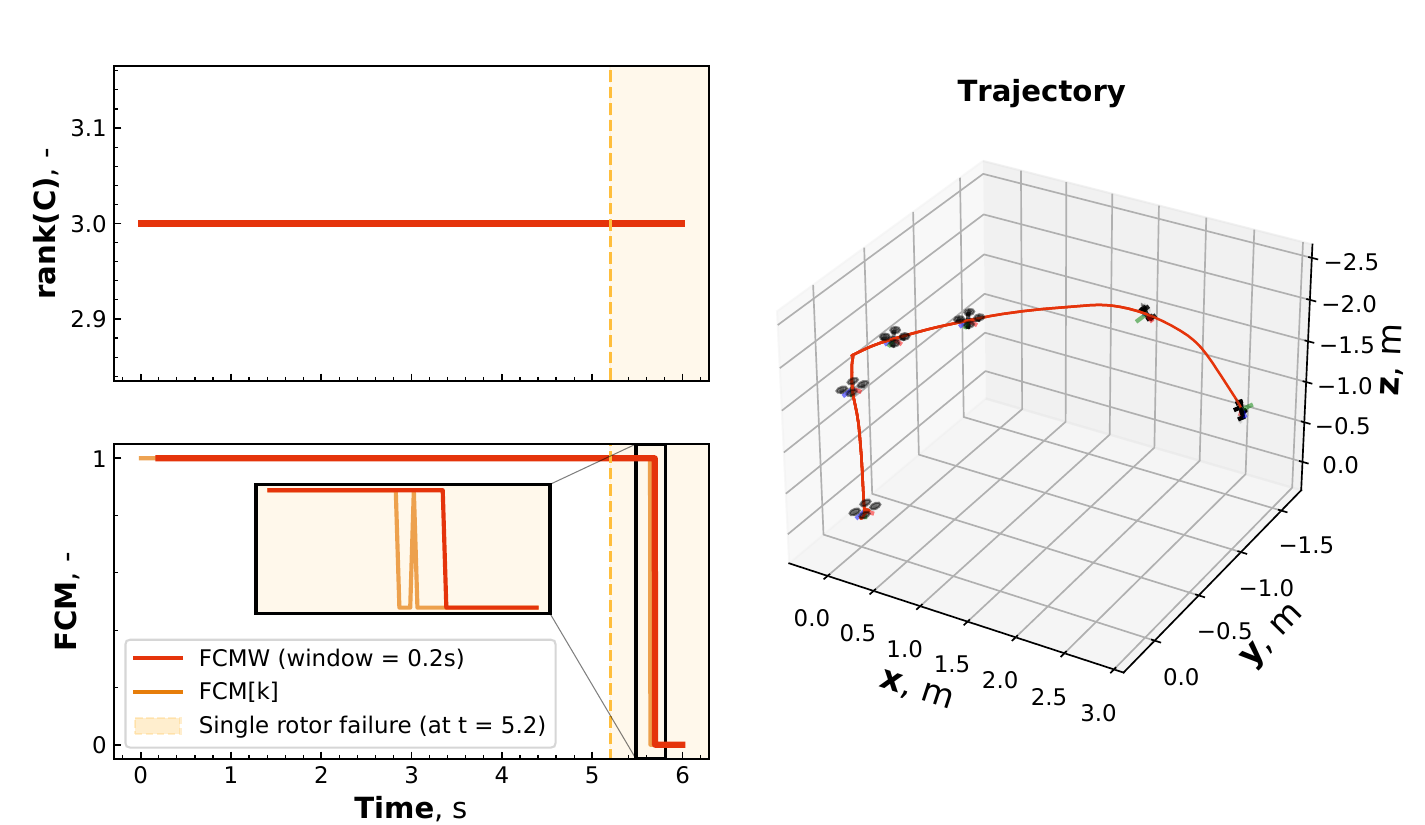}
    \caption{Position tracking trajectory of the quadrotor with the complete failure of a single rotor at $t=5.2$s. Also shown are the standard controllability metric, $rank(C) = n$, and those proposed in this paper, $FCM[k]$ and $FCMW$.}
    \label{fig:PosTrack_SRF}
\end{figure}

Conversely, when an actuator fault occurs during flight, the $FCM[k]$ and $FCMW$ are able to detect LOC shortly after the onset of the fault as shown in \cref{fig:PosTrack_SRF}. Crucially, this LOC event is not observable through the controllability check alone (i.e. $rank(C) = 3$ persists), even with the updated $B_{k+1} = B_{k}M$. This is because the elements of $C$ only approach zero, but do not reach it. This demonstrates the potential of the $FCM[k]$ and/or $FCMW$ for LOC detection. Furthermore, in \cref{fig:PosTrack_SRF}, the merits of using a voting window are highlighted by the more stable response of $FCMW$ compared to $FCM[k]$. However, this comes at the cost of a delayed LOC detection. Future work should seek to combine various window durations of $FCMW$ to refine estimates of LOC.



\subsection{Quadrotor Loss-of-control Flight Data Overview}
For a more practical demonstration, we here apply the $FCMW$ to real LOC flight data. As a means of comparison, the LOC flight data collected in the LOC prediction study of Altena et al. \cite{RNN_LOC_Altena} is used here. Flights are flown using a custom built quadrotor from commercial parts, the properties of which are summarized in \cref{tab:CineGo_Properties}. The flight data is collected in the CyberZoo - an open area of 7x7x10$m$ in height, width, and length respectively - at the Delft University of Technology. LOC is induced by demanding an excessively high yaw rate ($\pm 2000$ deg/s) from the quadrotor. Some time after the start of this manoeuvre, the quadrotor begins to oscillate about its roll and pitch axes with increasing magnitude until it makes a sudden, aggressive, turn culminating in a crash. More details on the quadrotors used, the LOC manoeuvre, and the flight data itself can be found in \cite{RNN_LOC_Altena}. Note that this LOC scenario is more nuanced than the clear-cut actuator fault case investigated in \cref{subsec:simRes}.

\begin{table}[!b]
\caption{Properties of the GEPRO CineGo quadrotor}
\label{tab:CineGo_Properties}
\centering
\begin{tabular}{lr} \hline
\multicolumn{1}{l}{\textbf{Mass (incl. batteries)}, $g$}          & 265           \\
\multicolumn{1}{l}{\textbf{Diagonal hub-to-hub}}    & 361           \\
\multicolumn{1}{l}{\textbf{  diameter}, $mm$}   & \\
\multicolumn{1}{l}{\textbf{Propeller diameter}, $mm$}              & 76.5          \\
\multicolumn{1}{l}{\textbf{Motor}, $-$}                           & Emax Eco 1407 3300kv            \\
\multicolumn{1}{l}{\textbf{Batteries}, $-$}                       & 1x Tattu R-Line 14.8V 550mAh 4S \\
\multicolumn{1}{l}{\textbf{Flight Controller (FC)}, $-$}          & MATEKSYS F722-mini 2-8S       \\
\multicolumn{1}{l}{\textbf{FC Software}, $-$}                     & Betaflight 4.2 \\ \hline
\end{tabular}
\end{table}

Moreover, in \cite{RNN_LOC_Altena}, LOC is defined as the moment in time where the roll or pitch angle exceeds, and continues to exceed, 90 deg after the start of the yawing manoeuvre. As stated by the authors, this definition is rather rudimentary and is tailored to the yaw LOC scenario. Indeed, the true moment of LOC may occur before, or after, the labelled moment. Therefore, the yaw manoeuvre flights which did not meet the LOC definition criteria of \cite{RNN_LOC_Altena} are also included in this study. Moreover, system identification flights for the CineGo quadrotor are also included in the data set as a ground truth for which LOC is not considered to occur. These include flights where a similar aggressive yaw manoeuvre is conducted, but stopped before any off-axis oscillations start. This culminates in a total of 121 flights: 12 are non-LOC flights and the remaining 109 harbor the dangerous - potentially LOC inducing - yaw manoeuvre, of which only 34 meet the attitude criteria of \cite{RNN_LOC_Altena}.

\subsection{Loss-of-control Detection on Real Flight Data}


From preliminary results, it was observed that applying a low-pass filter to the raw flight data (i.e. quadrotor rates and rotor speed measurements) improves the ability of the $FCMW$ to distinguish between the non-LOC and LOC flights. This is likely due to the noise present in the data, which results in false positives when the actuators are near, or at, saturation during an aggressive manoeuvre. Therefore, the $FCMW$ appears to be sensitive to noise in the data and adequate measures should be taken to minimize this effect. 

\begin{figure}[!b]
    \centering
    \includegraphics[width = \columnwidth]{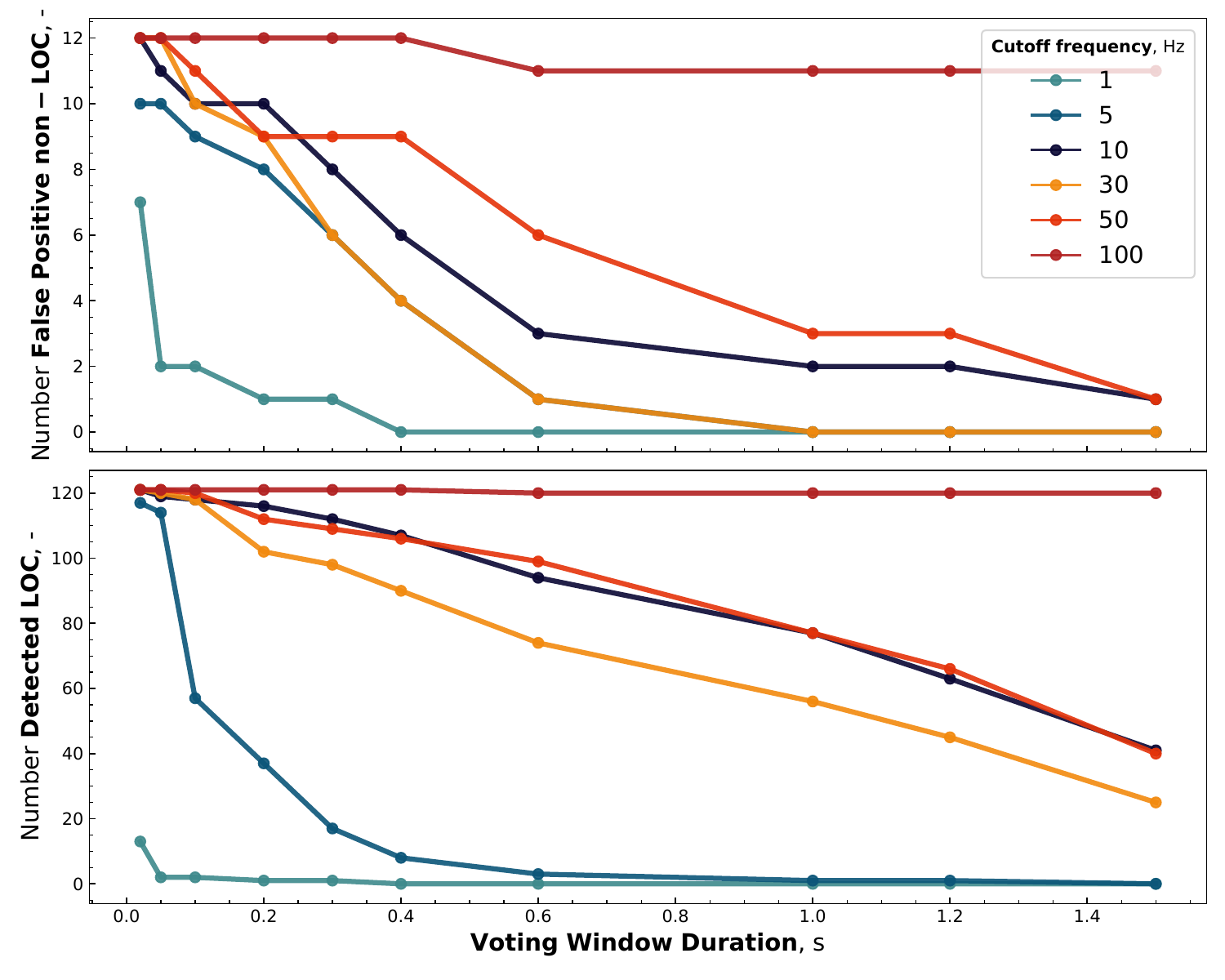}
    \caption{Performance of the Windowed Feasibly Controllable Metric ($FCMW$), applied to the CineGo loss-of-control flight data set \cite{RNN_LOC_Altena}, for various combinations of majority voting window, $MVW$, duration (in seconds) and low-pass filter cut-off frequency, $CF$ (in Hertz).}
    \label{fig:CineGo_Bonly_PerformanceSummary}
\end{figure}

Indeed, it is unclear what combination of majority voting window, $MVW$, and low-pass cut-off frequency, $CF$, yield the best LOC detection performance and how sensitive the $FCMW$ is to these hyper-parameters. Therefore, the $FCMW$ is computed across the whole flight data set for various combinations of voting window duration $MVW \in [0.02, 0.05, 0.1, 0.2, 0.4, 0.6, 1, 1.2, 1.5]$s and cut-off frequency $CF \in [1, 5, 10, 30, 50, 100]$Hz. The duration of the LOC inducing yaw manoeuvre lasts, on average, 2.4s before a crash, with the fastest crash occurring after 1.3s. Subsequently, the $MVW$ is capped at 1.5s to afford the $FCMW$ enough time to detect potential features which may occur within the period of the manoeuvre, but also to verify that windows longer than the LOC manoeuvre likely fail at detection altogether. Similar motivations for the cut-off frequency follow from power spectral densities of the raw flight data where the majority of the aerodynamics are observed to occur at frequencies below 50 Hz.


The ground truth non-LOC flights are used to determine which hyper-parameter combinations promote the greatest $FCMW$ sensitivity to LOC while limiting false positives of non-LOC. \Cref{fig:CineGo_Bonly_PerformanceSummary} summarizes the performance of the $FCMW$ across all the $MVW$ and $CF$ combinations. In general, the shorter the $MVW$, the more prone the $FCMW$ metric is to falsely label a non-LOC flight as a LOC flight. Longer $MVW$s reduce the sensitivity of the $FCMW$ (see top subplot of \cref{fig:CineGo_Bonly_PerformanceSummary}). Conversely, the lower the $CF$, the more capable the $FCMW$ is at correctly identifying a non-LOC flight. However, if the $CF$ is too low, then the $FCMW$ becomes insensitive to LOC detection as the majority of the aerodynamics are filtered out (refer to bottom subplot of \cref{fig:CineGo_Bonly_PerformanceSummary}). The optimum combination is found to be $MVW = 1.0$s and $CF = 30$Hz as this combination correctly identifies all non-LOC flights while being the most sensitive to LOC detection. As expected, this $MVW$ is below the fastest observed crash of 1.3s.  

\begin{figure}[!t]
    \centering
    \includegraphics[width = \columnwidth]{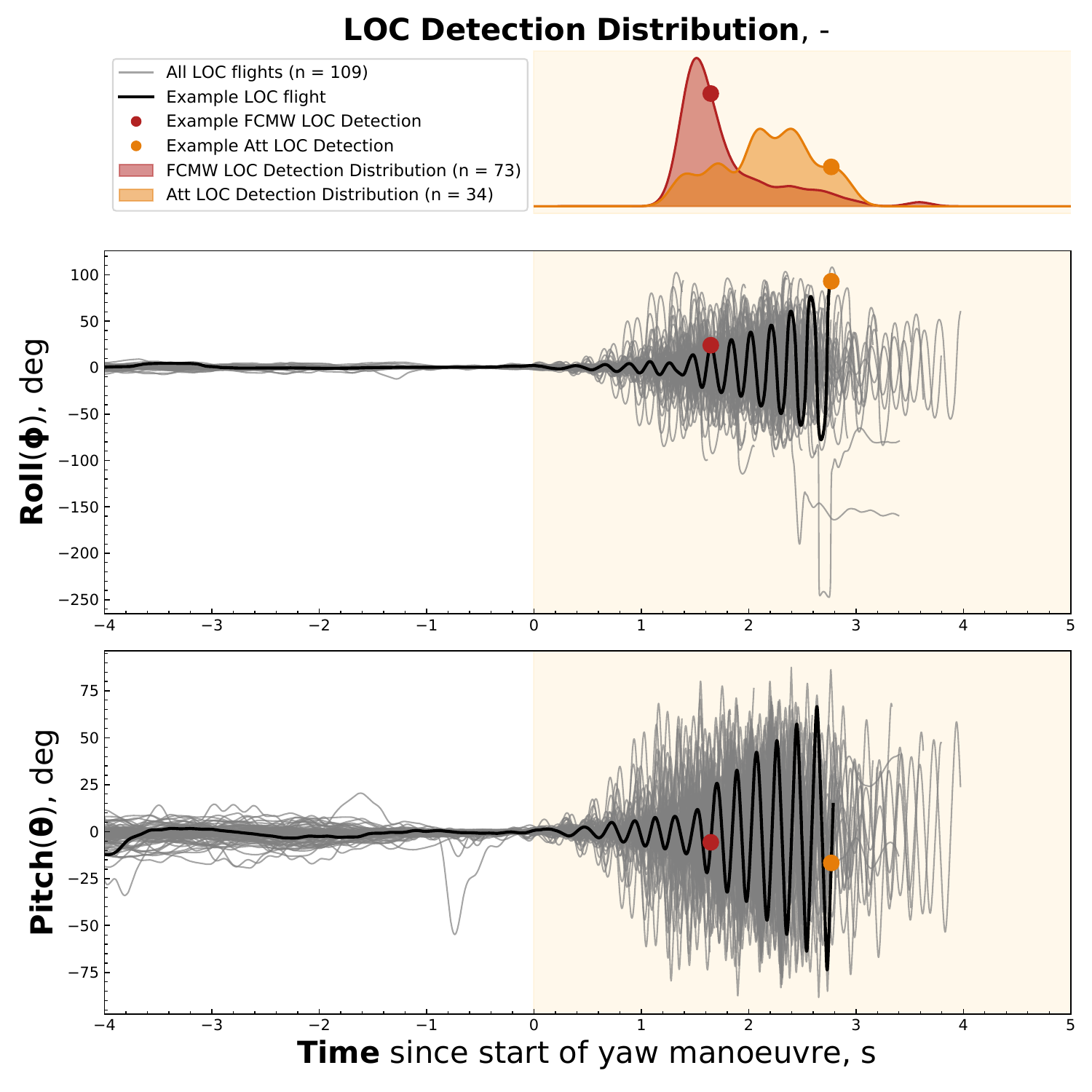}
    \caption{Comparison of the quadrotor loss-of-control (LOC) detection capabilities of the Windowed Feasibly Controllable Metric (FCMW) versus the attitude-based definition (Att) of \cite{RNN_LOC_Altena} for yaw-induced LOC flights of the CineGo quadrotor. Shown are the detection distributions of each definition. Plotted in black is an example LOC detection of both definitions for one of the LOC flights. The hyper-parameters for the FCMW are: voting window $MVW = 1.0$s and cut-off frequency $CF = 30$Hz.}
    \label{fig:FCMW_AllFlights_LOCDetection}
\end{figure}

\Cref{fig:FCMW_AllFlights_LOCDetection} illustrates the LOC detection capacity of the $FCMW$ detector, using the optimal hyper-parameters, for all yaw-LOC inducing flights of the CineGo. Also shown are the detection distributions of the $FCMW$ and attitude-based ($Att$) definition of \cite{RNN_LOC_Altena}, along with an example of these detectors applied to an arbitrary LOC flight plotted in black. It is clear that the $FCMW$ tends to detect LOC much earlier than the attitude definition, suggesting that LOC potentially occurs much earlier than initially expected in \cite{RNN_LOC_Altena}. Additionally, the $FCMW$ distribution suggests that this indicator is more consistent than its attitude-based counterpart. Interestingly, the $FCMW$ appears to detect LOC shortly after the unstable off-axis oscillations - which ultimately lead to the crash - begin. It is hypothesized that the $FCMW$ may be detecting the onset of a bifurcation, where the quadrotor transitions from a non-oscillatory equilibrium to an (unstable) oscillatory one. Indeed, the oscillatory response of the quadrotor may be reflective of the controller struggling to maintain control, or an increase in phase lag between the inputs and system response. These hypotheses should be verified in future experiments. 

\begin{figure}[!t]
    \centering
    \includegraphics[width = \columnwidth]{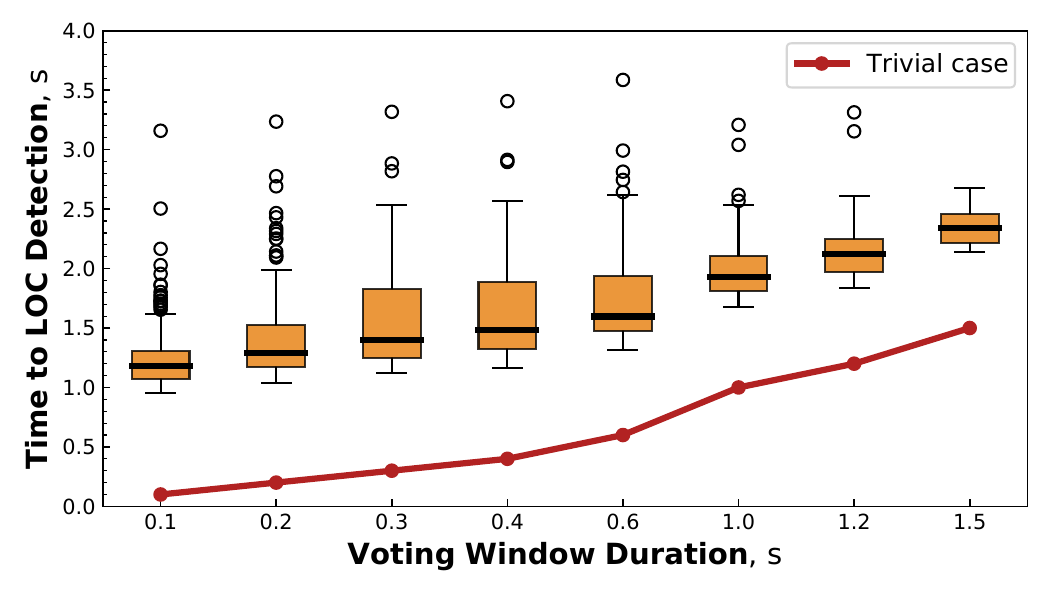}
    \caption{Variation in the time to loss-of-control detection of the Windowed Feasibly Controllable Metric ($FCMW$) as a function of the majority voting window ($MVW$) duration, in seconds. Also shown, in red, is the trivial case where the time-to-LOC detection = $MVW$. The two lowest $MVW$ (i.e. 0.02s and 0.05s) are not shown due to large variances and outliers.}
    \label{fig:MVW_Sensitivity_Bonly_CineGo}
\end{figure}

Another possibility is that the $FCMW$ simply flags LOC once the yaw manoeuvre lasts for the duration of the chosen voting window since the rotor speed commands saturate during the manoeuvre (i.e. the moment of LOC detection may be equivalent to the duration of $MVW$). This would imply that the $FCMW$ is not sensitive to LOC. To verify if this is the case, the time between the start of the yaw manoeuvre and LOC detection is evaluated for all $MVW$ durations at the optimal cut-off frequency of $CF = 30$ Hz (illustrated in \cref{fig:MVW_Sensitivity_Bonly_CineGo}). The majority of the LOC detections occur above the trivial case (i.e. time-to-LOC detection $\neq$ $MVW$), indicating that the $FCMW$ LOC detection is not only based on the duration of the $MVW$. While a larger $MVW$ does increase the time-to-LOC detection, this can be explained by the fact that more $FCM[k]$ need to be in the LOC state for the longer windows to trigger. In other words, the longer the $MVW$, the more severe the LOC event is. Moreover, recall that the non-LOC flights include the same aggressive yaw manoeuvres that induce LOC, only stopped before significant oscillations appear; none are detected as LOC by the $FCMW$ with the optimal hyper-parameters. This further suggests that the $FCMW$ is not simply flagging the yaw-manoeuvre itself, but rather other features indicative of LOC. As the $FCMW$ detects LOC earlier than the attitude definition of \cite{RNN_LOC_Altena}, it is perhaps unsurprising that it also labels more of the flights as LOC. Compare the 73 LOC detections of the $FCMW$ to the 34 of the attitude-based definition.

These experimental results show that - with appropriate hyper-parameters - the $FCMW$ is capable of more nuanced LOC detection for real, undamaged, quadrotor data. Moreover, the $FCMW$ does not flag nominal flights as LOC, even when similar manouevres are present. Taken in tandem with the simulation experiment results, the proposed metric ($FCMW$) appears to be a promising, and potentially general, indicator of (un)damaged quadrotor LOC.

\addtolength{\textheight}{-5cm}   
\bibliographystyle{ieeetr}  
\bibliography{root}



\end{document}